\documentclass[10pt,twocolumn,letterpaper]{article}

\usepackage{3dv}
\usepackage{times}
\usepackage{epsfig}
\usepackage{graphicx}
\usepackage{amsmath}
\usepackage{amssymb}

\usepackage{booktabs}
\usepackage{array}
\usepackage{multirow}
\usepackage{makecell}
\usepackage{gensymb}
\usepackage{diagbox}
\usepackage{hhline}
\usepackage[skip=2pt,font=small]{caption}

\usepackage[pagebackref=true,breaklinks=true,colorlinks,bookmarks=false]{hyperref}

\newcommand{\mat}[1]{\mathbf{#1}}

\threedvfinalcopy % *** Uncomment this line for the final submission

\def\threedvPaperID{293} % *** Enter the 3DV Paper ID here

\ifthreedvfinal\fi

\let\oldtwocolumn\twocolumn
\renewcommand\twocolumn[1][]{
	\oldtwocolumn[{#1}{
    	\begin{center}	\includegraphics[width=0.95\textwidth]{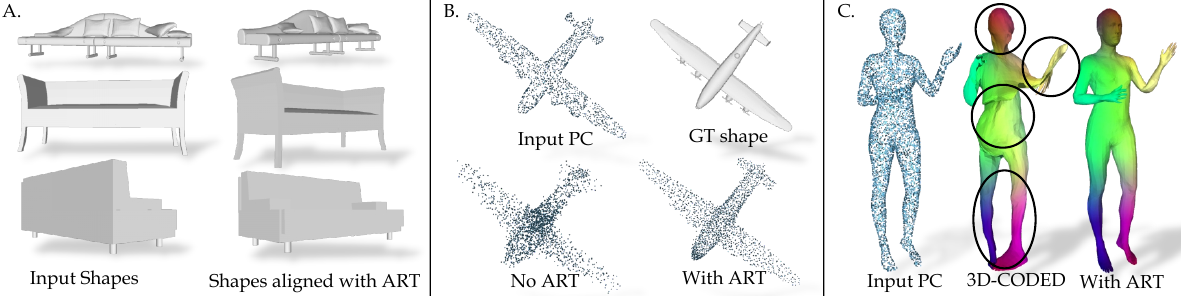}
            \captionof{figure}{Aligning 3D data is a hard problem which typically requires manual intervention, and most 3D learning SotA methods rely on good alignment.
            We propose Adjoint Rigid Transform (ART) Network, a self-supervised module which can be added to existing 3D networks to boost performance on a variety of tasks. Here we show (A) automatic shape alignment with ART; and use ART to improve (B) shape auto-encoding with PointNet~\cite{qi2017pointnet} and (C) human mesh registration with 3D-CODED~\cite{groueix2018b}.}
            \label{fig:teaser}
		\end{center}
	}]
}

\begin{document}

%%%%%%%%% TITLE
\title{Adjoint Rigid Transform Network: Task-conditioned Alignment of 3D Shapes}

\author{Keyang Zhou$^{1, 2}$ \quad Bharat Lal Bhatnagar$^{1, 2}$ \quad Bernt Schiele$^{2}$ \quad Gerard Pons-Moll$^{1, 2}$\vspace{0.1cm} \\
 $^1$University of Tübingen, Germany \\
 {\tt\small \{keyang.zhou,gerard.pons-moll\}@uni-tuebingen.de}\\
 $^2$Max Planck Institute for Informatics, Saarland Informatics Campus, Germany \\
 {\tt\small \{{bbhatnag, schiele\}@mpi-inf.mpg.de}}
}

\maketitle
% \thispagestyle{empty}

%%%%%%%%% ABSTRACT
\begin{abstract}
Most learning methods for 3D data suffer significant performance drops when the data is not carefully aligned to a canonical orientation. Aligning real world 3D data collected from different sources is non-trivial and requires manual intervention. 
In this paper, we propose the Adjoint Rigid Transform (ART) Network, a neural module which can be integrated with a variety of 3D networks to significantly boost their performance. ART learns to rotate input shapes to a learned canonical orientation, which is crucial for a lot of tasks such as shape reconstruction, interpolation, non-rigid registration, and latent disentanglement. ART achieves this with self-supervision and a rotation equivariance constraint on predicted rotations. With only self-supervision, ART facilitates learning a canonical orientation for both rigid and nonrigid shapes, which leads to a notable boost in performance of aforementioned tasks. Our code and model are available at~\cite{art}.
\end{abstract}

%%%%%%%%% BODY TEXT
\section{Introduction}
\begin{figure*}[t]
    \centering
    \includegraphics[width=\textwidth]{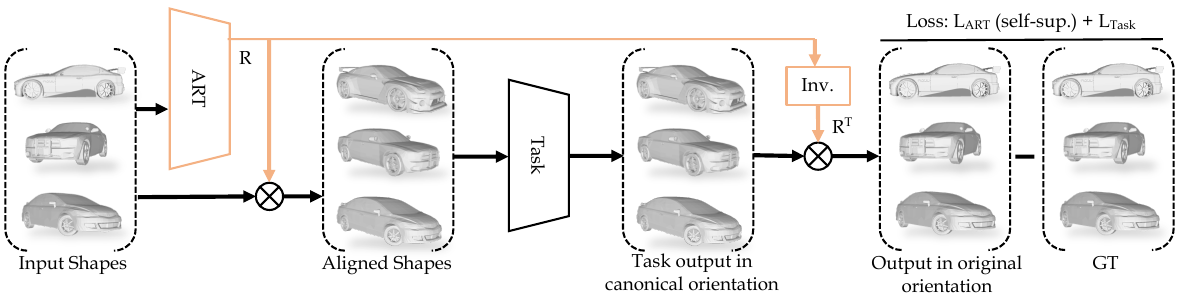}
    \caption{We propose a simple yet powerful module, Adjoint Rigid Transform (ART) Network that can automatically align 3D shapes to a common orientation. Given an input shape, ART predicts a rotation that maps the shape to a canonical orientation. The aligned shape is then fed to the target task and the output is rotated back to its original orientation. We show meshes here just for illustration purpose; the input can be either point cloud or mesh in practice.}
    \label{fig:overview}
\end{figure*}

With the rise of consumer grade 3D sensors and the popularity of a wide range of applications such as virtual and augmented reality, digital humans~\cite{SMPL:2015,fscape} and animals~\cite{zuffi20173d}, general objects~\cite{shapenet2015} and scenes~\cite{dai2017scannet}, there is an increasing demand for learning powerful representations from 3D data. 
Architectures such as PointNet~\cite{qi2017pointnet}, PointNet++~\cite{qi2017pointnet++}, or EdgeConv~\cite{wang2019dynamic} for processing point clouds, and Graph Convolutions~\cite{bouritsas2019neural,gong2019spiralnet++,kipf2017semi} for processing 3D meshes allow researchers to obtain impressive results on challenging tasks such as shape encoding~\cite{ranjan2018generating}, disentangling object shape and pose~\cite{zhou20unsupervised}, shape and pose interpolation~\cite{zhou20unsupervised} and 3D human mesh registration~\cite{groueix2018b,bhatnagar2020loopreg,bhatnagar2020ipnet}.
However, experiments show that the performance of these methods significantly decreases when 3D objects are not aligned to a common global orientation, which is a severe limitation since real world scanned objects are not aligned.

The question we pose here is: \emph{Can we automatically learn a network module to align shapes with only self-supervision?} The ability to do so would allow existing 3D networks to consume unaligned raw data and still keep a good performance.

The most common way to align shapes is with preprocessing, as done in ShapeNet~\cite{shapenet2015} and ModelNet~\cite{wu20153d}. But this is time-consuming and not completely automatic. The Spatial Transformer Networks (STN)~\cite{Jaderberg2015spatialtransformer} in 2D, and its 3D analogous incorporated in PointNet~\cite{qi2017pointnet}, allow the network to predict an affine transformation of the data to minimize the downstream loss. However, we experimentally show that STN alone does not achieve alignment. As a result, existing networks still struggle dealing with unaligned data.

We propose a simple yet powerful module, \emph{Adjoint Rigid Transform (ART)} Network, which can be plugged into 3D networks with an auto-encoder backbone and learn to automatically align 3D shapes conditioned on tasks including shape reconstruction, shape interpolation and nonrigid registration, as shown in Fig.~\ref{fig:teaser}. Different from previous work on learning rotation equivariance~\cite{esteves2018learning, shen20193d, thomas2018tensor}, ART does not depend on any specially designed feature or architecture, hence directly applicable to existing models.

An overview of our method is shown in Fig.~\ref{fig:overview}.
Our first key idea is to learn to rotate the object to a canonical orientation, perform the target task, and rotate it back to its original pose, see Fig.~\ref{fig:overview}.
This leads to canonical orientations which are convenient for the task, but many such orientations can arise during learning due to local minima, which hinders the network from achieving optimal performance.
To obtain a single canonical orientation per shape class, we show that the rotation predictor needs to be rotation equivariant.
Hence our second key idea is to impose a self-supervised equivariance loss during training. The remarkable result is that by just using self-supervision, ART can discover a single canonical orientation of general objects, which leads to a significant boost in task performance. 
This is desirable as downstream networks can focus on the task instead of devoting capacity to learning rotation equivariance.
Our contributions are as follows:
\begin{itemize}
    \item We propose a novel rotation equivariance loss, which allows ART to learn a canonical orientation conditioned on the target task. The alignment quality of ART is shown to be superior to previous works.
    \item We demonstrate that a wide range of 3D tasks lose performance when the data is not aligned. ART significantly improves the performance of existing methods on such tasks.
    \item We show the general applicability of ART spanning both rigid and non-rigid objects, and different input modalities such as meshes and point clouds.
\end{itemize}

\section{Related Work}
Existing methods lose performance for a wide range of 3D tasks when the data is not aligned to a canonical orientation.
In this section we first discuss works that address this issue by learning model features invariant or equivariant to global orientation. Next, we discuss works that explicitly align 3D data to a canonical orientation. Our method is related to both categories as ART can be added to existing methods and can align 3D shapes using self-supervision, thus making the downstream task more robust to varying global orientations.

\subsection{Rotation invariance/equivariance}
An interesting research direction to achieve rotation invariance is to handcraft descriptors that are invariant to rotation by design. These works are typically based on PCA~\cite{pca2003vranic} and geometric properties such as distances between pairs of points~\cite{2018Liu} (in 2D) or spherical harmonics~\cite{ALMAKADY2020102931} (in 3D).
A major limitation with the image-based descriptors is that they need to be manually designed per task which is non-trivial and difficult to scale and generalise to 3D tasks.
Another approach for learning rotation invariant representation is to bake rotation invariance into the network filters. Rotation invariance within a deep network can be obtained by explicitly rotating the feature maps or convolutional filters~\cite{Zhou2017ORN, pmlr-v48-dieleman16, steerable_cnn2017, NEURIPS2019_45d6637b, 2018Weiler, spezialetti2020learning}. Recently, Sun~\etal~\cite{sun2021canonical} proposed to learn a canonical frame by training with pairs of randomly rotated shapes. Our work follows a similar strategy, but we directly predict the canonical transformation from input shapes instead of using a capsule-based architecture. Hence ART can be seamlessly integrated into different shape representations and backbone models

\paragraph{Spatial Transformers}
Jaderberg \etal~\cite{jaderberg2015spatial} designed Spatial Transformer Network (STN), a differentiable module that manipulates the input image and feature maps by learned geometric transformations. Since it was first proposed, STN and its variants achieved promising results on image recognition~\cite{annunziata2018destnet}, image morphing~\cite{fish2020image} and image alignment~\cite{bas20173d, lin2017inverse}.

More related to our approach are works that apply STN to point clouds. PointNet~\cite{qi2017pointnet} inserted STN to both the input layer and feature maps as part of its architecture. ITN~\cite{yuan2018iterative} constrained the predicted transformations to be rotations and took an iterative procedure to gradually transform the input point clouds in small steps. Wang \etal~\cite{wang2019spatial} learned a combination of affine, projective and deformable transformations to dynamically update local patches for feature aggregation. Recently, Fang \etal~\cite{fang2020spatial} encoded spatial direction information of points using spatial transformers and defined an anisotropic filter for point clouds.

STN freely predicts affine transformations to warp the input and the feature maps. Although rotation is a subset of affine transformations, STN takes no measure to ensure that the network maps all the input data to a canonical orientation, which is crucial for many 3D tasks. ART on the other hand predicts rotations and explicitly encourages data to be aligned in a canonical orientation, hence promoting better performance of downstream tasks. 

\subsection{3D data alignment}
Aligning data to a consistent frame of reference is a crucial preprocessing step for many applications~\cite{villena2020deep}.
When the correspondences between two shapes are available, the rigid transformation relating them can be analytically computed~\cite{arun1987least}. But in real world scenarios where correspondences are often unknown, Iterative Closest Points (ICP)~\cite{chen1992object, besl1992method} and its variants~\cite{pomerleau2015review} are more widely used.
Since the iterative procedure of ICP starts from an initial guess of transformation, it can easily get trapped in local minima due to bad initialization.
Recently, Liu~\etal \cite{uncorr2020liu} proposed an approach to learn unsupervised correspondences which can be used as an alternative to ICP.

For aligning a collection of shapes, Principal Component Analysis (PCA) offers a naïve solution by matching the principal axes of every shape to those of the reference shape. Although aligning with PCA alone is error-prone, PCA is the basis for many alignment algorithms~\cite{sfikas2011rosy+, kazhdan2004shape}. For symmetric objects, reflection symmetry can also be used to further improve the alignment~\cite{podolak2006planar, chaouch2008novel, minovic1993symmetry}.
Huang \etal~\cite{huang2013fine} discretized the transformation sampling space for each shape and formulated joint alignment as a Markov Random Field optimization problem. Averkiou \etal~\cite{averkiou2016autocorrelation} observed that ambiguities in alignment often arise in only a few candidate orientations. They reduced the search space and evaluated candidate alignments by analyzing the auto-correlation descriptors of shapes.
We compare ART with a PCA baseline and Averkiou \etal~\cite{averkiou2016autocorrelation}. We show that our method outperforms them both.

\section{Adjoint Rigid Transform Network}
\begin{figure}[t]
    \centering
    \includegraphics[width=\linewidth]{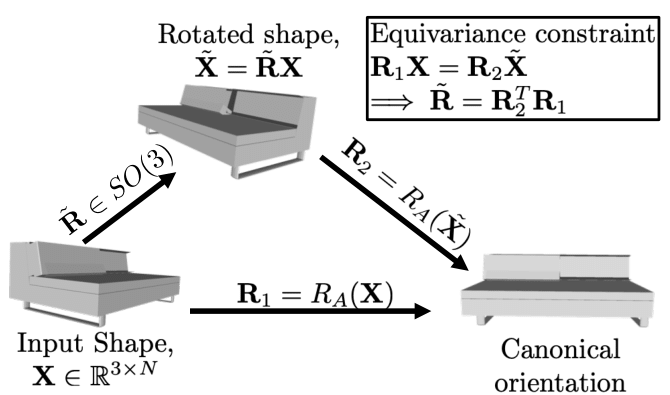}
    \caption{Equivariance constraint for ART. Given an input shape $\mat{X} \in \mathbb{R}^{3\times N}$, we sample a random rotation $\tilde{\mat{R}}$ and obtain the rotated shape $\tilde{\mat{X}}=\tilde{\mat{R}}\mat{X}$. Our equivariance constraint forces the network $R_A$ to map
    $\mat{X}$ and $\tilde{\mat{X}}$ to the same canonical space.
    }
    \label{fig:equivariance_constraint}
\end{figure}

In this section, we describe the general formulation of Adjoint Rigid Transform (ART) Network and how it integrates with the target task. We also introduce the equivariance loss which encourages the learned canonical orientation to be unique. 

\subsection{Canonical Rotation Prediction}
Consider a function $g:\mathcal{X} \mapsto \mathcal{Y}$ which takes a matrix of points $\mat{X} \in \mathcal{X}$ as input, and let $\mathbf{R} \in \mathrm{SO}(3)$ be a rotation matrix. 
It is often useful to obtain $g$ as a composition of functions

\begin{equation}
g(\mat{X}) = \mathbf{R}^T g^\prime (\mat{R} \mat{X}).
\label{eq:adjoint}
\end{equation}
The idea is to rotate the input shape to a canonical orientation where processing $g^\prime(\cdot)$ is more natural, and then rotate the output back with $\mat{R}^T$. This is what we refer to as an adjoint transform, as is popularly used in robotics~\cite{murray1994mathematical}.

The adjoint transform in Eq.~\ref{eq:adjoint} inspires the design of ART. In perception tasks however, objects can appear in multiple orientations, and the canonical coordinate frame is not defined a priori.
Our task thus boils down to predicting a rotation $\mat{R}$ which aligns object $\mat{X}$ to a canonical view without using any pose labeling.
Suppose we have $\mat{X}\in \mathbb{R}^{3\times N}$ as input. $\mat{X}$ contains unordered point coordinates for point cloud input or ordered vertex coordinates for mesh input.
We confine the predicted transformation to rotation for the following considerations: i) Orientation is often the most notable factor of variation in unprocessed 3D data.
ii) Rotation preserves the intrinsic geometry of the shape which is desirable for a number of tasks. iii) Unlike general affine transformations, rotation is guaranteed to be invertible. Rotations can be conveniently inverted by taking the transpose. This choice ensures the well-definedness and efficiency of adjoint transformations.

Hence, we learn a mapping $R_A(\mat{X}):\mathcal{X}\mapsto \mathrm{SO}(3)$ from the input shape $\mat{X}$ to the canonical rotation $\mat{R}$.
We denote the downstream 3D network by $g^\prime$ and denote the network with ART module as a whole by $g$. 
With the predicted rotation $R_A(\mat{X})$, Eq.~\ref{eq:adjoint} can be formulated as:
\begin{equation}
    g(\mat{X}) = R_A^T(\mat{X})g^\prime(R_A(\mat{X})\mat{X})
    \label{eq:adjoint_learnt}
\end{equation}
which replaces the known rotation in Eq.~\ref{eq:adjoint} with a learned one.
Note that when the predicted rotation is identity $R_A = \mat{I}_{3\times3}$, we are left with a standard downstream network.

ART can be trained end-to-end with downstream tasks where the expected output orientation is the same as input (see Fig.~\ref{fig:overview}). Shape auto-encoding is a typical use case for ART. In this case, the desired output is the input shape itself, which can be imposed with the following modified self-reconstruction loss 
\begin{equation}
 L_\mathrm{recon} = d\left(\mat{X} , R_A^T(\mat{X})g^\prime(R_A(\mat{X})\mat{X})\right),
 \label{eq:recon_loss}
\end{equation}
where the loss function $d: \mathbb{R}^{3 \times{N}} \times \mathbb{R}^{3 \times{M}} \mapsto \mathbb{R}$ can be Chamfer distance for point clouds or vertex-to-vertex distance for meshes.

\subsection{Rotation Equivariance}
Predicting rotations as in Eq.~\ref{eq:adjoint_learnt} with an objective function from Eq.~\ref{eq:recon_loss} does not ensure alignment. ART can learn multiple canonical orientations for a shape collection and easily get trapped in local minima.
This is undesirable for both shape alignment and the target task. A necessary condition for learning unique canonical orientation is that the same shape in different input orientations should be transformed to the same canonical orientation, mathematically $R_A(\mat{X})\mat{X}=R_A(\mat{R}\mat{X})\mat{R}\mat{X} \;\forall \mat{R} \in \mathrm{SO}(3)$.
From this it follows that the rotation predictor $R_A$ should be equivariant to input orientations, \ie $R_A(\mat{R}\mat{X}) = R_A(\mat{X})\mat{R}^T$.
We enforce this constraint using a rotation equivariance loss (Fig.~\ref{fig:equivariance_constraint}).

During training, given an input $\mat{X} \in \mathbb{R}^{3\times N}$, we uniformly sample a rotation $\mat{\tilde{R}} \in \mathrm{SO}(3)$ and obtain a different orientation of $\mat{X}$, $\mat{\tilde X} = \mat{\tilde{R}}\mat{X}$. Suppose we have $\mat{R}_1 = R_A(\mat{X})$ and $\mat{R}_2 = R_A(\mat{\tilde{X}})$. Then from the unique orientation constraint $\mat{R}_1\mat{X} = \mat{R}_2{\mat{\tilde X}}$ we can derive
\begin{equation}
    \mat{\tilde R} = \mat{R}_2^T\mat{R}_1.
    \label{rot_equi}
\end{equation}
Since $\mat{\tilde R}$ is known, we can turn Eq.~\ref{rot_equi} into a loss term
\begin{equation}
L_{\mathrm{rot\_matrix}}=\left\| \mat{\tilde{R}}-\mat{R}_2^T\mat{R}_1\right\| _{2}^{2}.
\label{rot_equi_loss}
\end{equation}
Note that $L_{\mathrm{rot\_matrix}}$ is imposed on rotation matrices. However, objects such as tables can have rotational symmetry. Suppose the rotation symmetry group of $\mat{X}$ is of order $n$. Then there are $n$ possible rotation matrices that will leave shape $\mat{X}$ unchanged, so there is not a well-defined groundtruth. To handle the potential ambiguity in rotational symmetry, we add another loss term:
\begin{equation}
L_{\mathrm{rot\_chamfer}}=d_{\mathrm{CD}}\left(\mat{\tilde{X}},\mat{R_2}^T\mat{R_1X}\right),
\end{equation}
where $d_{\mathrm{CD}}$ is the symmetric Chamfer distance. The intuition behind this loss term is that shapes remain invariant to distinct elements of their symmetry groups.

Combining these two terms, we have
\begin{equation}
    L_{\mathrm{ART}} = \lambda_1L_{\mathrm{rot\_matrix}} + \lambda_2L_{\mathrm{rot\_chamfer}}.
\end{equation}

\subsection{Implementation Details}
When the input shapes reside on the same plane, sampling one rotation $\tilde R$ per iteration suffices for Eq.~\ref{rot_equi_loss}. For general $\mathrm{SO}(3)$ rotations, we sample three $\tilde R$ instead. We also took the approach of~\cite{yuan2018iterative} and iteratively apply ART to refine rotation predictions.

We adopt different architectures for Adjoint Rigid Transform Network according to different types of input. For point clouds, we use PointNet~\cite{qi2017pointnet} as backbone. PointNet is more efficient in both training and inference compared with more sophisticated architectures for point cloud processing such as PointNet++~\cite{qi2017pointnet++}, adding minimal overhead to the downstream network. For meshes, we simply use mesh down-sampling layers~\cite{ranjan2018generating} and fully-connected layers. We use the continuous rotation representation proposed in~\cite{Zhou_2019_CVPR} for rotation prediction. Details about the architecture are in the supplementary.

We need to ensure that ART does not disturb training when the dataset is already in alignment, so we initialize it to predict the identity matrix. In the following experiments, we set $\lambda_1=0.02$, and set $\lambda_2=0.05$ for table category and $\lambda_2=0$ otherwise. Inputs are centered and normalized to fit within a unit ball. 
\section{Experiments}

\begin{figure*}[t]
    \centering
    \includegraphics[width=\textwidth]{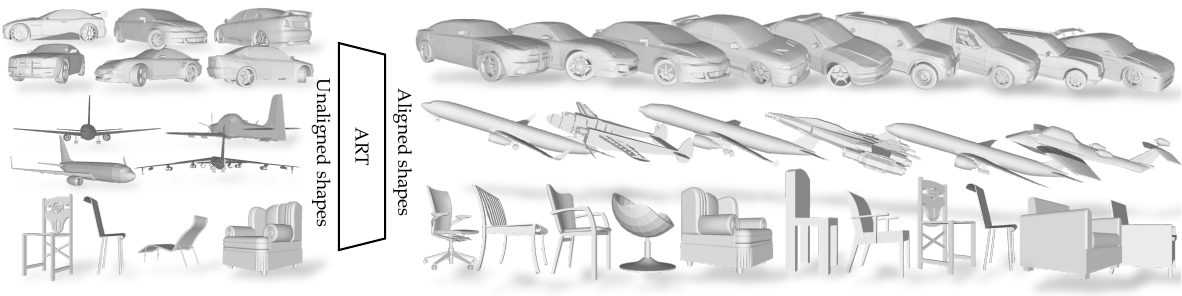}
    \caption{
    ART can align 3D shapes with just self-supervision. Given input shapes in arbitrary orientations (left), ART can align them to a common orientation (right).}
    \label{fig:shapenet_alignment}
\end{figure*}

\begin{figure*}[t]
    \centering
    \includegraphics[width=\textwidth]{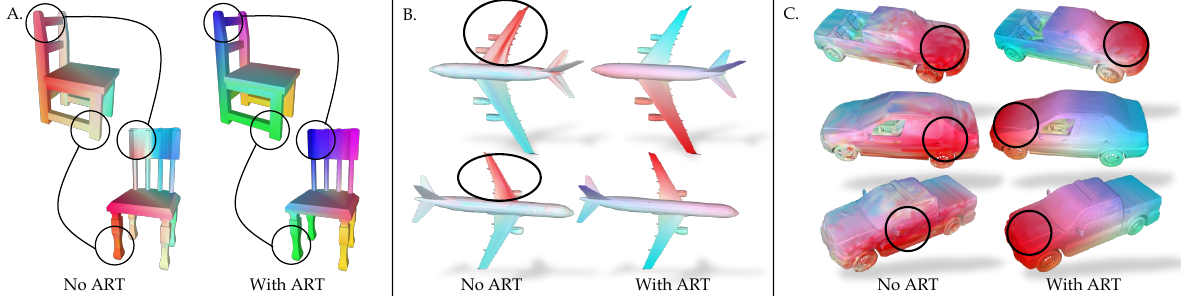}
    \caption{
    ART can establish correspondences between different objects within a ShapeNet category even when the shapes are not aligned. In A, we show that baseline~\cite{achlioptas2018learning} does not assign same colours to the corresponding parts. In B, we show that \cite{achlioptas2018learning} gets confused between the left and the right wing whereas ART does not. In C, we can see that \cite{achlioptas2018learning} cannot find consistent correspondences as the front of the car is mapped to the rear and side of the car respectively. ART on the other hand can establish correct correspondences.}
    \label{fig:shapenet_correspondences}
\end{figure*}

\begin{figure*}[t]
    \centering
    \includegraphics[width=\textwidth]{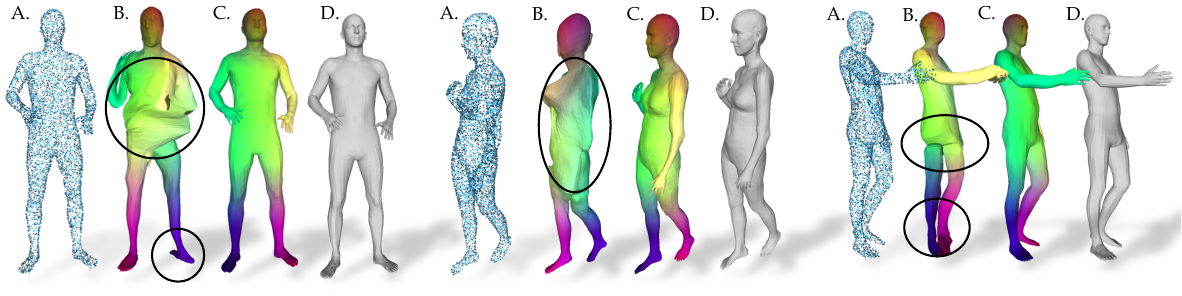}
    \caption{ART improves 3D human registration as compared to 3D-CODED~\cite{groueix2018b} (single init.). In each set we show, (A) input point cloud, (B) registration with \cite{groueix2018b}, (C) registration with \cite{groueix2018b}+ART (Our) and (D) GT mesh. Adding ART improves the performance of \cite{groueix2018b}.}
    \label{fig:human_registration}
\end{figure*}

\begin{figure*}
    \hspace{0.02\linewidth}
    \begin{minipage}[t]{0.32\linewidth}
    \includegraphics[width=0.9\linewidth]{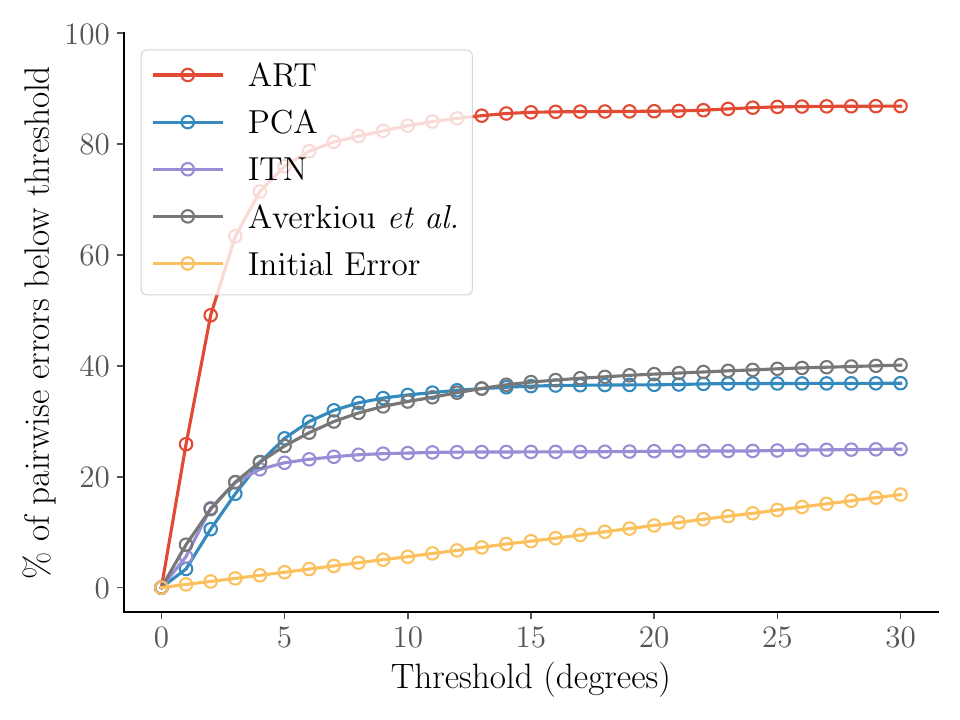}
    \caption{Percentage of shape pairs with an angular distance less than the given thresholds. ART outperforms the other three baselines by a large margin, with around 80\% of shape pairs differing by less than $10\degree$.}
    \label{fig:angle_dist}
    \end{minipage}
    \hspace{0.02\linewidth}
    \begin{minipage}[t]{0.32\linewidth}
    \includegraphics[width=\linewidth]{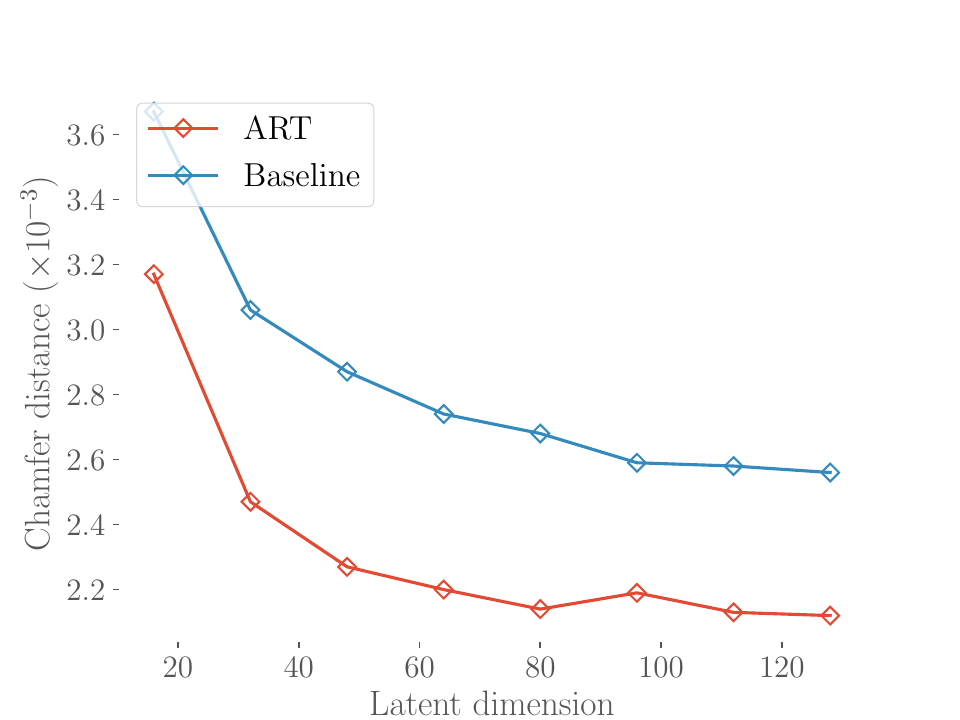}
  \caption{Reconstruction error (mean Chamfer distance) vs. size of latent code for ART and \cite{achlioptas2018learning}. It can be seen that with a latent code of size 32, ART already outperforms ~\cite{achlioptas2018learning}  with a latent code of size 128.
}
    \label{fig:latent_size}
    \end{minipage}
    \hspace{0.02\linewidth}
    \begin{minipage}[t]{0.32\linewidth}
    \includegraphics[width=0.9\linewidth]{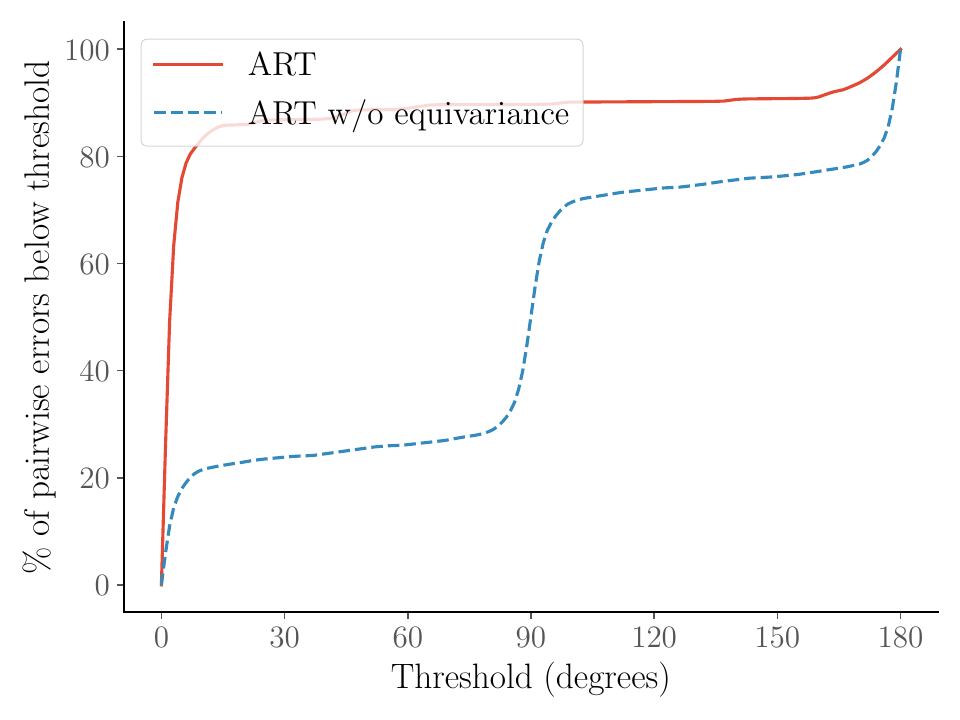}
    \caption{ART w/o equivariance constraint can learn multiple canonical orientations. Notice the sharp rise of the blue curve at $90\degree$ and $180\degree$ thresholds. It indicates the four modes of orientations on the plane category.}
    \label{fig:angle_dist_ablation}
    \end{minipage}
\end{figure*}

In this section, we evaluate the effectiveness of ART through extensive experiments.
We demonstrate the usefulness of ART on several tasks for 3D data involving rigid (ShapeNet~\cite{shapenet2015}) and non-rigid (humans~\cite{AMASS:ICCV:2019}) objects.
We also show that ART works seamlessly across different shape representations such as point clouds and meshes.

\subsection{Point Cloud Auto-encoding}
\begin{figure}[t]
    \centering
    \includegraphics[width=\linewidth]{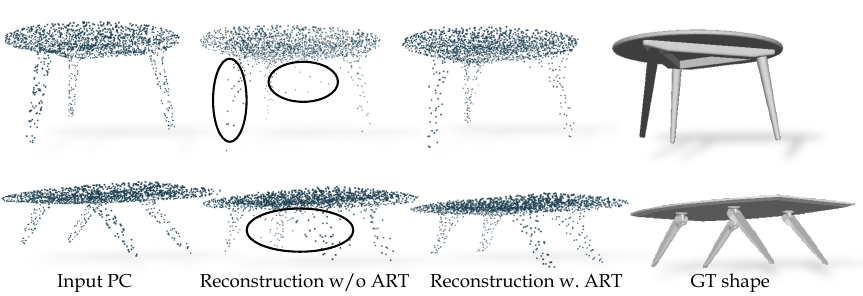}
    \caption{Shape auto-reconstruction. ART improves reconstruction quality when ShapeNet~\cite{shapenet2015} objects are not aligned.}
    \label{fig:shapenet_auto_recon}
\end{figure}

Shape auto-encoding is a fundamental task in unsupervised feature learning of point clouds. Prior works typically report performance on datasets consisting of axis-aligned shapes such as ShapeNet~\cite{shapenet2015}. But in the real world, shapes are often collected from diverse sources and are therefore unaligned. To assess the capability of ART to handle unaligned data, we perturb shapes in ShapeNetCore~\cite{shapenet2015} by applying random 2D rotations around the gravity axis as well as random 3D rotations. We evaluate auto-encoding in both single-category and multi-category settings. For single-category auto-encoding, we use plane, car, chair, table and sofa categories. For multi-category auto-encoding, we jointly train on these categories. For each shape, we uniformly sample 8000 points. The training/validation/testing splits are kept as 85\%/5\%/10\%. We use symmetric Chamfer distance as both training loss and evaluation metric.

We use the network from Achlioptas \etal~\cite{achlioptas2018learning} as the auto-encoder backbone. It also serves as one of our baselines. On top of it, we add ITN~\cite{yuan2018iterative} and ART to assess gains in performance brought by each module. Moreover, we designed a PCA baseline, where we naïvely align shapes by matching their principal axes and train the plain auto-encoder on PCA-aligned data.

Table~\ref{tab:pc_ae} reports the numerical results for point cloud auto-reconstruction on both aligned and unaligned data. We can observe a significant increase in reconstruction error when the auto-encoder trains on unaligned data. Pre-aligning shapes with PCA alleviates the problem, but there is still a performance gap. ITN performs on par with PCA, while ART outperforms all baselines on unaligned data. When the training data is perturbed by a 2D azimuthal rotation, ART matches or even beats the performance on pre-aligned data for most categories. This performance gain drops slightly for the case of 3D rotations. Qualitative examples are shown in Fig.~\ref{fig:shapenet_auto_recon}.

\begin{table*}[t]
\centering
\begin{tabular}{| >{}p{0.13\textwidth} | >{\centering}p{0.11\textwidth} | 
>{\centering}p{0.08\textwidth}  >{\centering}p{0.08\textwidth}  >{\centering}p{0.08\textwidth}  >{\centering}p{0.08\textwidth}  >{\centering}p{0.08\textwidth} | c |}
\hline
\multirow{2}{*}{\textbf{Method}} & \multirow{2}{*}{\textbf{Data}} & \multicolumn{5}{c|}{\textbf{Single-category}} & \multirow{2}{*}{\textbf{Multi-categories}} \\
& & Plane & Chair & Car & Table & Sofa &  \\
\hline
a) AE~\cite{achlioptas2018learning} & pre-aligned & 1.16 & 2.20 & {1.81} & 2.37 & 2.21 & {2.02} \\
\hhline{|=|======|=|}
b) AE & unaligned 2D & 1.86 & 3.20 & 2.39 & 2.96 & 3.14 & 2.54 \\
\hline
c) $+$PCA & unaligned 2D & 1.34 & 2.42 & 1.91 & 3.14 & 2.41 & 2.26 \\
\hline
d) $+$ITN~\cite{yuan2018iterative} & unaligned 2D & {1.37} & {2.29} & {1.89} & { 2.38} & {2.20} & {2.10} \\
\hline
e) $+$ART (Our) & unaligned 2D & {\bf 1.16} & {\bf 2.15} & {\bf 1.87} & {\bf 2.32} & {\bf 2.13} & {\bf 2.03} \\
\hline
\hhline{|=|======|=|}

f) AE & unaligned 3D & 3.31 & 3.83 & 3.37 & 5.18 & 3.84 & 3.53 \\
\hline
g) $+$PCA & unaligned 3D & 1.38 & 2.44 & 1.97 & 3.46 & 2.61 & 2.41 \\
\hline
h) $+$ITN & unaligned 3D & {1.37} & {2.55} & {1.89} & { 2.85} & {2.46} & {2.43} \\
\hline
i) $+$ART (Our) & unaligned 3D & {\bf 1.22} & {\bf 2.41} & {\bf 1.88} & {\bf 2.38} & {\bf 2.16} & {\bf 2.26} \\

\hline
\end{tabular}
\caption{We evaluate our approach on point cloud auto-encoding by comparing to multiple baselines in different settings. Numbers are reported in Chamfer distances ($\times 10^{-3}$). We show that performance of existing method~\cite{achlioptas2018learning} drops significantly between aligned and unaligned data. Aligning shapes with PCA and ITN~\cite{yuan2018iterative} improves performance but our method clearly outperforms all the baselines, often even matching the \emph{oracle} performance on pre-aligned data. We highlight the lowest error in each training setting with boldface.}
\label{tab:pc_ae}
\end{table*}

\subsection{Shape Alignment}
One of the key advantages of our method is alignment of shapes to a canonical orientation (see Fig.~\ref{fig:shapenet_alignment}).
We evaluate alignment on ShapeNet~\cite{shapenet2015} plane category, which we perturbed by applying random azimuthal rotations. The plane category has a well-defined criteria for exact alignment. It also doesn't suffer from ambiguities of rotational symmetry, making it suitable for benchmarking alignment accuracy. We add results on other categories in supplementary.

Here, we use the alignment evaluation metric proposed by Averkiou \etal~\cite{averkiou2016autocorrelation}. We compute the angular distance between every pair of shapes using ShapeNet groundtruth orientation as reference.
We compare ART to PCA, ITN~\cite{yuan2018iterative}, and the alignment method proposed by Averkiou \etal~\cite{averkiou2016autocorrelation} which assumes consistent up vectors among shapes.

The cumulative distribution curve for pairwise errors is shown in Fig.~\ref{fig:angle_dist}. For completeness we also include the initial pairwise error for unaligned shapes. We can see that Averkiou \etal~\cite{averkiou2016autocorrelation} perform on par with PCA on this dataset while ITN underperforms, with over half of the shape pairs differing by more than $30\degree$. This is because all these methods get confused by the near-symmetries of planes (e.g. matching plane tip to tail, or wings to fuselage). ART has two advantages in this regard, i) the downstream auto-encoding task allows ART to learn semantically meaningful features to better disambiguate parts of the plane, and ii) our equivariance constraint forces ART to chose a unique canonical orientation for all planes. More qualitative results on 2D and 3D alignment will be shown in supplementary.

\subsection{Shape Interpolation}
\begin{figure*}[t]
    \centering
    \includegraphics[width=\textwidth]{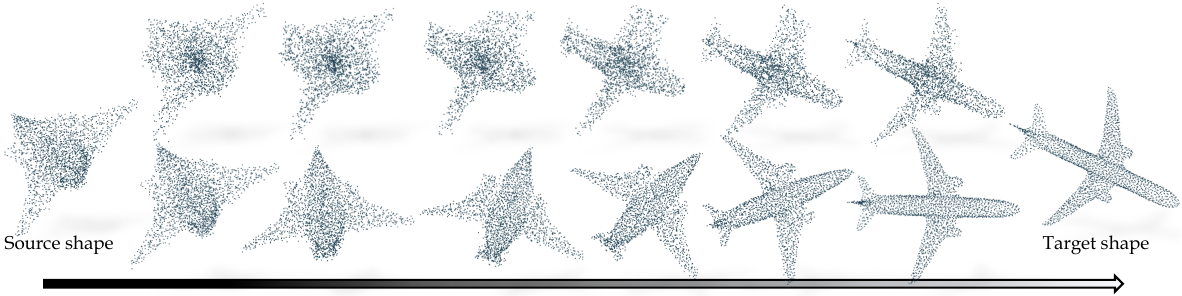}
    \caption{Shape interpolation \emph{without} ART (top) leads to severe distortions due to different orientations between source and target shapes. +ART (bottom) on the other hand aligns the source and the target shape to a canonical orientation, resulting in smooth interpolation.}
    \label{fig:shapenet_interp}
\end{figure*}

Shape interpolation is a challenging task for unaligned shapes as linear interpolation in feature space cannot handle the highly nonlinear global orientation. In Fig.~\ref{fig:shapenet_interp}, we show that  interpolation with~\cite{achlioptas2018learning} severely distorts the shape when the source and target shapes have different orientations. Notably, ART does not suffer from this as it brings both the source shape and target shape to the same orientation where interpolation becomes meaningful. 

\subsection{Shape Correspondence Prediction}
We show that our method can be used to predict dense correspondences between objects within a ShapeNet~\cite{shapenet2015} category. The input to our shape auto-encoder is an unordered point cloud of an object in arbitrary orientation. It predicts a fixed number of points as output, which approximates the input point cloud. This allows us to establish correspondences across shapes as each shape is represented as deformations of the same fixed set of points. We compare the correspondences predicted by \cite{achlioptas2018learning} with and without ART. Notice that the model doesn't have access to supervision of correspondence, but point-wise correspondences naturally arise when adding ART since the output space of the auto-encoder now has a consistent global orientation.
The qualitative results are shown in Fig.~\ref{fig:shapenet_correspondences}. 

\subsection{Human Body Registration}
3D-CODED, proposed by Groueix \etal~\cite{groueix2018b}, is a popular learning-based human mesh registration approach. Given an unordered point cloud as input, 3D-CODED learns to deform a pre-defined human body template mesh to match the point cloud. Since 3D-CODED was trained on synthetic shapes with consistent global orientations, it can only reconstruct shapes in that particular orientation. When dealing with real world scans without alignment, 3D-CODED applies multiple initializations to find the optimal orientation -- it rotates the scan with $\sim$100 uniformly sampled rotations around the gravity axis.

We show that adding ART to their method solves this problem. ART learns a canonical orientation for humans during training. At inference time, it only takes a single forward pass to transform the input to the canonical orientation, which is both faster and more accurate than sampling rotations. We compare the performance of ART with 3D-CODED~\cite{groueix2018b} in both single-initialization and multiple-initialization cases. We test on registered meshes from Renderpeople~\cite{renderpeople} and AMASS~\cite{AMASS:ICCV:2019}. We can see from Table~\ref{tab:3dcoded} that ART consistently outperforms 3D-CODED by a large margin, even though we do not require 100 initializations.
The qualitative results are shown in Fig.~\ref{fig:human_registration}.

\begin{table}[t]
\centering
\begin{tabular}{|l|c|c|}
\hline
\diagbox[width=.22\textwidth]{Method}{Dataset} & Renderpeople & AMASS \\
\hline
3D-CODED Single Init. & 193.0 & 50.0\\
\hline
3D-CODED Multi Init. & 23.8& 33.9 \\
\hline
ART (Our) & \textbf{16.9}& \textbf{16.8}\\
\hline
\end{tabular}
\caption{Human mesh registration. We report vertex-to-vertex error in mm. 3D-CODED+ART with single initialization outperforms 3D-CODED~\cite{groueix2018b} with multiple ($\sim$100) initializations.}
\label{tab:3dcoded}
\end{table}

\subsection{Human Pose Transfer}
\begin{figure*}[t]
    \centering
    \includegraphics[width=\linewidth]{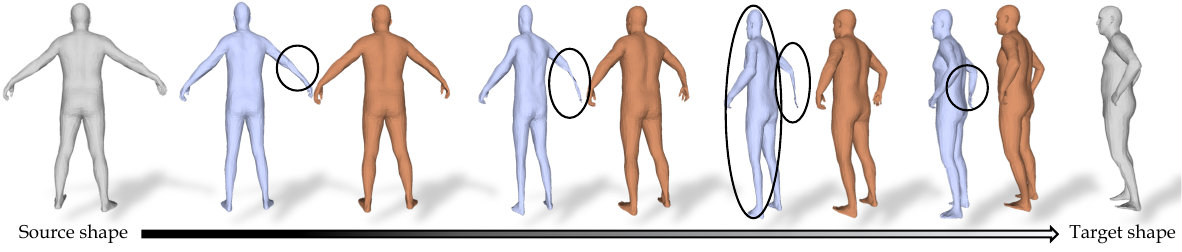}
    \caption{Pose interpolation \emph{without} ART (blue) leads to squeezing artifacts whereas +ART (brown) handles global rotations well. We use Zhou \etal~\cite{zhou20unsupervised} for pose interpolation and compare performance with and without adding ART to the method.}
    \label{fig:pose_interp}
\end{figure*}

\begin{table}[t]
\centering
\begin{tabular}{|c|c|c|}
\hline
Method & Supervised & Unsupervised \\
\hline
Zhou \etal~\cite{zhou20unsupervised} & 15.44 & 19.43 \\
\hline
ART (Our) & {\bf 9.16} & {\bf 17.98} \\
\hline
\end{tabular}
\caption{ART for human pose transfer. ART improves the performance of ~\cite{zhou20unsupervised} in both supervised and unsupervised setups. We report vertex-to-vertex error in mm. }
\label{tab:art_transfer}
\end{table}

We use the SotA unsupervised model proposed by Zhou \etal~\cite{zhou20unsupervised} for the task of pose transfer.
The input to their method is a registered 3D human mesh and they decompose the mesh deformations into shape and pose components. Like~\cite{zhou20unsupervised}, we train on AMASS~\cite{AMASS:ICCV:2019}, a human motion capture dataset parametrized by SMPL model~\cite{SMPL:2015}. We evaluate model performance on the task of pose transfer, where we reconstruct a human body from the shape of one subject and the pose from another subject. Since we have access to the underlying SMPL parameters, we utilize SMPL model to generate pseudo-groundtruth for evaluation.
Following the practice of~\cite{zhou20unsupervised}, we also trained a supervised model from SMPL pseudo-groundtruth. Table~\ref{tab:art_transfer} summarizes pose transfer errors. ART significantly lowers the pose transfer error in both supervised and unsupervised setups.

\subsection{Human Pose Interpolation}
Another limitation of~\cite{zhou20unsupervised} is that it cannot interpolate poses between two humans with very different global orientations.  Fig.~\ref{fig:pose_interp} shows pose interpolation results when there is a large global rotation between source and target. In~\cite{zhou20unsupervised}, pose interpolation is done by linearly interpolating between source and target pose codes. However, the intermediate pose codes do not always lie on the pose manifold, causing the squeezing artifact. Since ART can explicitly factor global rotations out, pose codes in our approach only represent articulation and not global rotation. To interpolate pose, we simply apply linear interpolation to pose codes, and spherical linear interpolation~\cite{shoemake1985animating} to global rotations predicted by ART.
Fig.~\ref{fig:pose_interp} clearly demonstrates the strength of our method. See supplementary for more results.

\subsection{Analysis and Ablation}

\paragraph{Size of latent code vs. performance} We show that a point cloud auto-encoder~\cite{achlioptas2018learning} cannot learn to encode shape orientation despite using a large latent dimension. In this experiment, we gradually increase the size of the latent code and study the performance improvement in the task of multi-category point cloud reconstruction. It can be seen from Fig.~\ref{fig:latent_size} that while the error drops with an increasing latent dimension, the gap between the baseline and ART remains. ART with 32 latent dimensions already outperforms~\cite{achlioptas2018learning} with 128 latent dimensions.

\paragraph{Importance of the equivariance constraint}
We implement a baseline where ART is free to predict arbitrary rotations, \ie we do not enforce equivariance constraint. It can be seen from Fig.~\ref{fig:angle_dist_ablation} that the quality of alignment without the equivariance constraint is poor as the network is free to pick multiple canonical orientations without the constraint.

\section{Conclusion}
Learning 3D representations is a challenging task and methods across a wide range of tasks rely on aligned data. Obtaining this alignment in real world scenarios is not trivial and often requires a lot of manual effort. We propose a simple module, \emph{Adjoint Rigid Transform (ART) Network} that can automatically align 3D data conditioned on a set of target tasks. What makes ART effective is the semantic features learned by the auto-encoder coupled with a rotation equivariance constraint which results in a canonical orientation for input shapes. ART can be easily integrated into existing systems, can work with point clouds and meshes, and can be trained with self-supervision. We experimentally show that ART significantly boosts the performance of existing methods for shape auto-encoding, alignment, and interpolation on rigid objects, and human registration, pose transfer, and interpolation on non-rigid humans.

\vspace{0.3cm}
\noindent
\textbf{Acknowledgements} 
This work is supported by the German Federal Ministry of Education and Research (BMBF): Tübingen AI Center, FKZ: 01IS18039A. This work is funded by the Deutsche Forschungsgemeinschaft (DFG, German Research Foundation) - 409792180 (Emmy Noether Programme, project: Real Virtual Humans). Gerard Pons-Moll is a member of the Machine Learning Cluster of Excellence, EXC number 2064/1 – Project number 390727645. The project was made possible by funding from the Carl Zeiss Foundation.

{\small
\bibliographystyle{ieee_fullname}
\bibliography{egbib}
}

\end{document}